\documentclass[10pt,twocolumn,letterpaper]{article}

\usepackage[pagenumbers]{wacv} 

\usepackage{graphicx}
\usepackage{amsmath}
\usepackage{amssymb}
\usepackage{booktabs}
\usepackage{multirow}
\usepackage[table]{xcolor}
\usepackage[accsupp]{axessibility}  

\usepackage[pagebackref,breaklinks,colorlinks]{hyperref}

\usepackage[capitalize]{cleveref}
\crefname{section}{Sec.}{Secs.}
\Crefname{section}{Section}{Sections}
\Crefname{table}{Table}{Tables}
\crefname{table}{Tab.}{Tabs.}

\def\wacvPaperID{1039} 
\def\confName{WACV}
\def\confYear{2024}

\begin{document}

\title{Learning to Detour: Shortcut Mitigating Augmentation \\ for Weakly Supervised Semantic Segmentation}

\author{JuneHyoung Kwon\textsuperscript{1}, Eunju Lee\textsuperscript{2}, Yunsung Cho\textsuperscript{2}, YoungBin Kim\textsuperscript{1,2}\\
\textsuperscript{1}Department of Artificial Intelligence, Chung-Ang University, Korea \\
\textsuperscript{2}Graduate School of Advanced Imaging Science, Multimedia \& Film, Chung-Ang University, Korea \\
{\tt\small \{dirchdmltnv,dmswn5829,cho4062002,ybkim85\}@cau.ac.kr}
}
\maketitle

\begin{abstract}
Weakly supervised semantic segmentation (WSSS) employing weak forms of labels has been actively studied to alleviate the annotation cost of acquiring pixel-level labels. However, classifiers trained on biased datasets tend to exploit shortcut features and make predictions based on spurious correlations between certain backgrounds and objects, leading to a poor generalization performance. In this paper, we propose shortcut mitigating augmentation (SMA) for WSSS, which generates synthetic representations of object-background combinations not seen in the training data to reduce the use of shortcut features. Our approach disentangles the object-relevant and background features. We then shuffle and combine the disentangled representations to create synthetic features of diverse object-background combinations. SMA-trained classifier depends less on contexts and focuses more on the target object when making predictions. In addition, we analyzed the behavior of the classifier on shortcut usage after applying our augmentation using an attribution method-based metric. The proposed method achieved the improved performance of semantic segmentation result on PASCAL VOC 2012 and MS COCO 2014 datasets. 
\end{abstract}

\section{Introduction}

Acquiring pixel-level labels for semantic segmentation is both cost- and time-intensive. To alleviate this problem, weakly supervised semantic segmentation (WSSS) is actively researched by different groups. WSSS uses a weak label that contains less information about the location of an object than a pixel-level label but has a cheaper annotation cost. Examples of such weaker forms of labels are image-level class labels\cite{lee2021reducing, lee2021anti, lee2022threshold}, bounding boxes\cite{khoreva2017simple, lee2021bbam, song2019box}, points\cite{bearman2016s, kim2022beyond}, and scribbles\cite{tang2018normalized, lin2016scribblesup}. Among these weak labels, we focus on the image-level class label, which is the most accessible and has the lowest annotation cost.

\begin{figure}[t]
\begin{center}
   \includegraphics[width=1\linewidth]{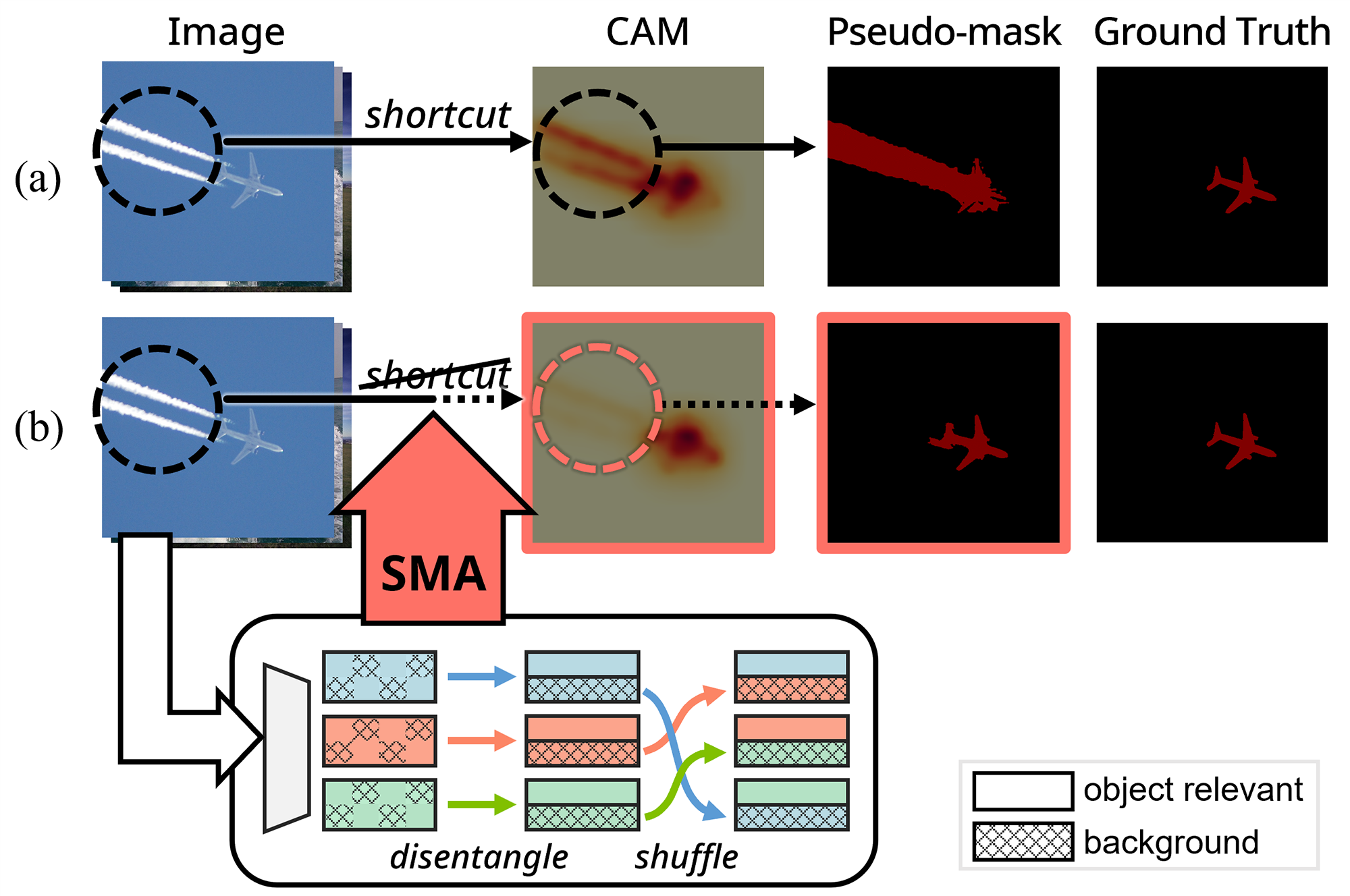}
\end{center}
\vspace*{-4mm}
   \caption{(a) The classifier trained with CutMix utilizes background cues as shortcuts producing suboptimal localization map. (b) Our proposed SMA relies less on contexts and focuses more on objects to generate fine-grained localization maps.}
\label{fig:fig1}
\end{figure}

In most studies on WSSS utilizing image-level class labels, class activation map (CAM) is used as the initial seed which can be seen as estimation of the region covered by the object. The classifier is trained to predict the corresponding category of the image and identify the target object region. However, the classifier unexpectedly highlights the background region, which in turn generates a blurry CAM\cite{lin2016scribblesup}. This is because the classifier exploits the bias of the dataset as a shortcut instead of making predictions using the information related to the target object. Here, a ``shortcut'' refers to a decision rule that performs well in typical contexts but fails in slightly different circumstances\cite{geirhos2020shortcut}. This background bias originates from the biased dataset composed of images in which a specific object frequently co-occurs with a certain background context\cite{geirhos2020shortcut}. For instance, an object corresponding to the ``sheep'' class almost always appears in ``grass landscape'' and the visual layout is similar in many images. Among the Pascal VOC 2012 training datasets, more than 20\% of the images with ``sheep'' class contain ``grass landscape'' as the context. The same trend is observed for cow–grass landscape, boat–water, and aeroplane–sky train–track pairs\cite{lee2022weakly}. Therefore, a classifier trained with a biased dataset depends not only on the target object but also leverages the background bias as a shortcut to make predictions. Because the classifier benefits from this non-generalizable shortcut feature, it occasionally assigns high class scores on the background regions, and fails to activate the target object region, where such objects appear outside typical scenes.

Several studies have been conducted on data augmentation\cite{devries2017improved, zhang2017mixup, yun2019cutmix} that enables the training of classifiers with enhanced robustness\cite{cao2022survey}. These methods minimize the effect of biases on the classifier and aid in the identification of non-discriminative regions of the target object while making predictions. However, these approaches are susceptible to dataset bias as they do not take into account the contextual information of the image during the augmentation of the training samples. Figure~\ref{fig:fig1} (a) illustrates the localization map produced by applying a conventional augmentation. The CutMix\cite{yun2019cutmix}-trained classifier leverages ``sky'' region (especially contrail part), which often appears together with object ``aeroplane'', as a shortcut, resulting in an inaccurate CAM and the pseudo-mask.

In this paper, we propose \textbf{Shortcut Mitigating Augmentation (SMA)}, which prevents the use of shortcuts in the classifiers, for WSSS. We reduce the shortcut usage degree by presenting the classifier with synthetic features of the object–background combinations that do not frequently appear in the training dataset. First, we disentangle the object-relevant and background features, because these features are highly entangled with each other and mislead the classifier into identifying background as objects of interest. Each feature is exclusively aggregated, and contrastive learning is adopted for further seperation. Second, based on the these representations, we randomly shuffle the background representation in each mini-batch, while the foreground representation is fixed, and vice versa. Through the proposed SMA, the classifier receives representations of various object-background combinations, allowing it to focus more on the target object without being biased towards a specific background. As a result, the dependence of the classifier on the shortcuts is reduced, and a high-quality CAM is obtained. 

We performed an additional experiment to evaluate the extent to which each augmentation method relied on shortcuts. To analyze the classifier's shortcut behavior, we introduced a metric based on attribution methods. The experimental results showed that compared to the existing augmentation method, the classifier trained with SMA employed more object-relevant information and less background information when identifying the target objects. By leveraging the shortcut mitigation effect of SMA, the mean Intersection over Union (mIoU) of the localization map and pseudo-mask were enhanced on the Pascal VOC 2012 and MS COCO 2014 datasets. 

The main contributions of this study are as follows: We \textbf{(1)} proposed SMA, a data augmentation method for WSSS that synthesize the novel representation of various object-background combinations \textbf{(2)} evaluated the effectiveness of our method on reducing shortcut usage by adopting attribution method-based metric \textbf{(3)} enhanced the performance of the localization map and pseudo-mask by applying the SMA method on various WSSS methods using the Pascal VOC 2012 and MS COCO 2014 benchmarks for the WSSS. 

\section{Related work}

\subsection{Weakly supervised semantic segmentation} 

WSSS methods that use image-level class labels generate a localization map based on the initial seed CAM, and then produce a pseudo-mask through an additional refinement process. Because the initial seed identifies only the discriminative regions in the image, numerous studies have been conducted to expand such regions. AdvCAM\cite{lee2021anti} identifies more regions of objects by manipulating the attribute map through adversarial climbing of the class scores. DRS\cite{kim2021discriminative} suppresses the most discriminative region, thereby enabling the classifier to capture even the non-discriminative regions. SEAM\cite{wang2020self} regularizes the classifier so that the differently transformed localization maps are equivalent. AMN\cite{lee2022threshold} leverages a less discriminative part through per-pixel classification. 

Further, several studies have been conducted to develop feasible methods to prevent the classifier from learning misleading correlations between the target object and the background. SIPE\cite{chen2022self} captures the object more accurately through prototype modeling of the background. ICD\cite{fan2020learning} includes an intra-class discriminator that discriminates the foreground and background within the same class. W-OoD\cite{lee2022weakly} utilizes out-of-distribution data as extra supervision to train the classifier to suppress spurious cues. $C^2$AM\cite{xie2022c2am} generate class-agnostic activation maps by adopting foreground-background contrastive learning. While $C^2$AM leverages separated representation to refine CAM, SMA synthesizes the disentangled representation to diversify the views learned by the classifier. In addition, various studies have employed a saliency map as an additional supervision or for post-processing\cite{lee2021railroad, fan2020cian, lee2019ficklenet, wei2017object, wei2018revisiting, yao2020saliency}. In contrast, our proposed method disentangles the background information in the representation space, and thus, no additional supervision is required.

\subsection{Data augmentation}

Data augmentation improves the generalization ability of a classifier for unseen data by improving the diversity of the training data. The image erasing method removes one or more sub-regions in an image and replaces them with zero or random values. Cutout\cite{devries2017improved} randomly masks a specific part of the image, and Hide-and-Seek\cite{singh2018hide} allows the classifier to identify the class relevant features after randomly hiding the patch in the image. In contrast, the image mix-based method mixes two or more images. Mixup\cite{zhang2017mixup} interpolates two images and labels, and CutMix\cite{yun2019cutmix} replaces a certain region of an image with a patch of another image. However, because the method that uses this regional patch randomly occludes the sub-regions, including both the object and background areas, the classifier trained with this method cannot distinguish the foreground from the context. In addition, there are methods to augment the representation in the feature space\cite{lim2021noisy, verma2019manifold}. Furthermore, there have been similar approaches to enhance the diversity of the training dataset by combining features of different instances\cite{zhou2021mixstyle, kim2023wedge}. SMA differs in its approach by explicitly disentangling two features and bidirectionally shuffling them.

In context decoupling augmentation (CDA)\cite{su2021context}, the copy-and-paste method is introduced to a WSSS task. CDA decouples the object and context by pasting the pseudo-mask obtained in the first stage to another image. However, this method uses a single class image to obtain an accurate object instance and restricts the scale of the mask. Consequently, the diversity of the augmented representation is limited. Our proposed method synthesizes features irrespective of the number or size of the object mask, and thus, it can provide a more diverse representation to the classifier.

\section{Method}

In section \ref{SMA}, we introduce the SMA method, which generates synthetic features that encapsulate combinations of objects and backgrounds that do not frequently occur together. The degree of shortcut usage of the SMA and other augmentation methods is analyzed in section \ref{analysis}. Finally, section \ref{schemes} describes additional training schemes for performing an effective augmentation. 

\subsection{Shortcut mitigating augmentation}
\label{SMA}

Here, we describe the overall procedure of the SMA. The proposed method disentangles the input features into object-relevant and background features, and then shuffles different training samples to generate synthetic features. A classifier trained with SMA minimizes the use of background cues as shortcuts and leverages more target objects to create higher-level localization maps. 

\begin{figure}[t]
\begin{center}
   \includegraphics[width=1\linewidth]{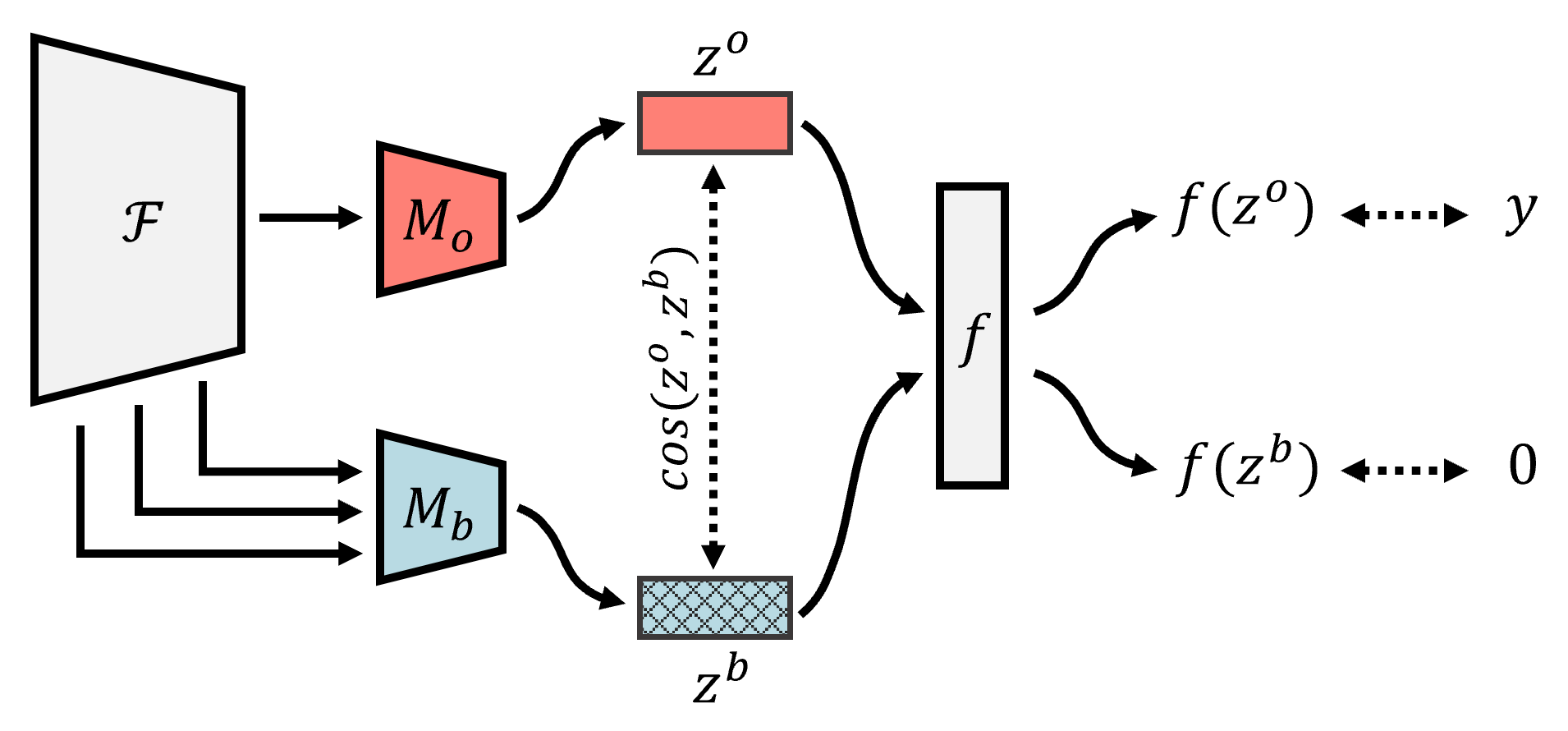}
\end{center}
\vspace*{-5mm}
   \caption{An illustration demonstrating how the classifier separates object-relevant features from background features. The two aggregators $M_o$ and $M_b$ disentangle the features obtained from the backbone network $\mathcal{F}$ into object-relevant feature $z^o$ and background feature $z^b$. Then two classification score $f(z^o), f(z^b)$ obtained by classification head $f(\cdot)$ is supervised by label $y$ and $0$ respectively. In order to further differentiate $z^o, z^b$, contrastive loss is additionally employed.}
\label{fig:fig2}
\end{figure}

\noindent \textbf{Disentangling object-relevant and background features.} First, we disentangle the features obtained through the backbone network into object-relevant and background features. To separate distinctive features effectively, mutually exclusive attributes must be aggregated from the input feature. Global average pooling (GAP), which is generally used for pooling, is not appropriate as it can include peripheral attributes. Therefore, inspired by previous studies\cite{lee2021reducing, araslanov2020single, zhu2021background}, we aggregated the object-relevant and background features using different aggregators $M_o, M_b$ respectively. In addition, to integrate the background related information more effectively, we extract features from shallow layers of the backbone network, since semantics of background are not restricted\cite{chen2022self}. We fed extracted features to $M_b$. The distinct features are aggregated by $M_o, M_b$ through the following process.

Given $N$ images $\{x_i\}_{i=1}^N$, the feature map $\{z_i\}_{i=1}^N$ is obtained through the backbone network $\mathcal{F(\cdot)}$, in which $z_i \in \mathbb{R}^{C \times h \times w}$. $C, h$, and $w$ indicate the channel size, height, and width of the feature map, respectively. Based on the extracted feature map $z_i$, different $1\times1$ convolutions $\phi(\cdot), \theta(\cdot)$ are employed, and $\mathsf{Softmax}$ operation is performed on the spatial dimension to produce the attention maps $o_i \in \mathbb{R}^{d \times h \times w}, b_i \in \mathbb{R}^{d \times h \times w}$. The attention maps $o_i, b_i$ can be seen as spatial importance for and object-relevant and background regions, respectively. $o_i, b_i$, and $z_i$ are flattened, i.e., $o_i \in \mathbb{R}^{d \times hw}, b_i \in \mathbb{R}^{d \times hw}$, and $z_i \in \mathbb{R}^{C \times hw}$. Then, the disentangled representation $z_i^o, z_i^b$ can be calculated as follows:

\begin{equation}
z_i^o = m(o_i \otimes z_i^T), z_i^b = m(b_i \otimes z_i^T)
\label{eq:equation1}
\end{equation}

\begin{figure*}[t]
\begin{center}
   \includegraphics[width=0.85\linewidth]{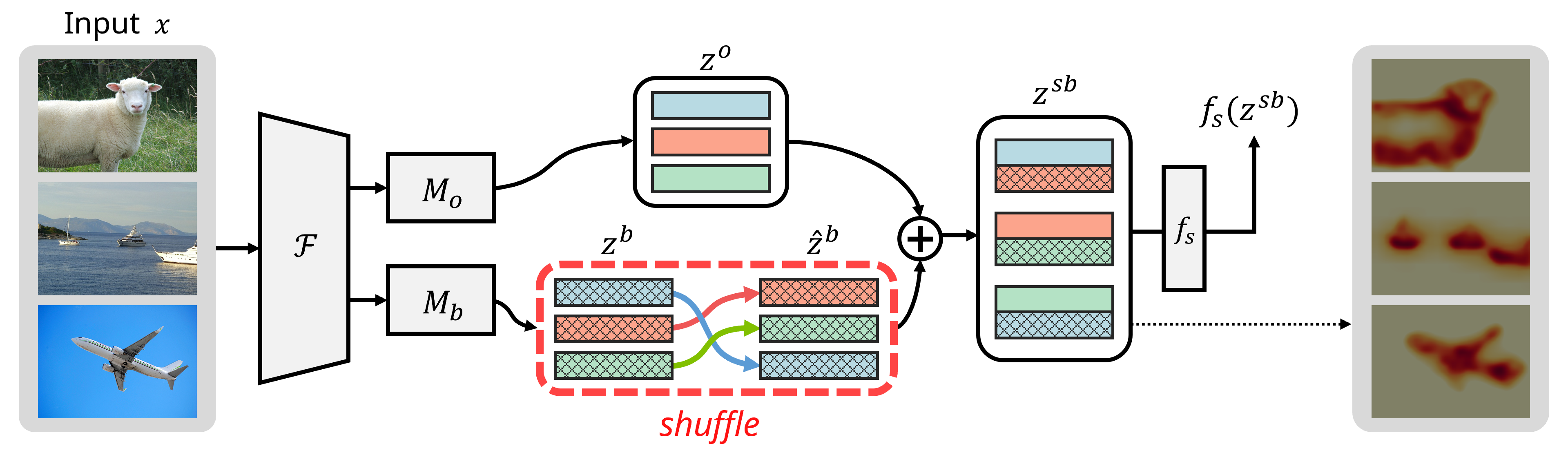}
\end{center}
\vspace*{-5mm}
   \caption{This is an overview of SMA when only background feature shuffling is implemented. The feature map $z$ is disentangled into object-relevant representation $z^o$ and background representation $z^b$ by aggregators $M_o$ and $M_b$. We perform background shuffling to acquire $\hat{z}^b$ in which the background representation is permuted on the mini-batch. Then novel representation $z^{sb}$ is synthesized by concating $z^o$ and $\hat{z}^b$. $f_s({z^{sb}})$ denotes prediction by $f_{s}$ and $\oplus$ denotes channel-wise concatenation operation.}
\label{fig:fig3}
\end{figure*}

where $\otimes$ and $m(\cdot)$ denote the matrix multiplication and average operations over the channel dimension, respectively; $z_i^o \in \mathbb{R}^{1 \times C}, z_i^b \in \mathbb{R}^{1 \times C}$ are the object-relevant and background representations, respectively. We effectively separate the disentangled representations by supervising them with different labels for each logit value obtained using the classification head $f(\cdot)$. We supervise logit value $f(z^o)$ with class label $y$, so that $z^o$ contains object relevant attributes. Following \cite{lee2022weakly, zhu2021background}, we assign zero vector label $y=(0, 0, \cdots, 0)$ to score $f(z^b)$ to ensure $z^b$ only includes background attributes. Further, binary cross entropy ($BCE$) is used for training the classifier, and the classification loss is determined as follows:

\begin{equation}
L_{cls} = BCE(f(z_i^o), y) + BCE(f(z_i^b), 0)
\label{eq:equation2}
\end{equation}

We adopt contrastive loss to separate distinct representations further. Our training objective forces the object-relevant representation $z^o$ and background representation $z^b$ apart. The contrastive loss used in this case is calculated as follows:

\begin{equation}
L_{contr} = - {{1} \over {N}} \sum_{i=1}^N \log(1 - sim(z_i^o, z_i^b))
\label{eq:equation3}
\end{equation}

where $sim(z_i^o, z_i^b)$ denotes the cosine similarity between the two representations; $z_i^o$ and $z_i^b$ are disentangled such that the distance between them on the feature space increases through the contrastive loss. The overall process of feature disentanglement can be seen in Figure~\ref{fig:fig2}. 

\noindent \textbf{Shuffling disentangled representations.} We randomly shuffle the disentangled representations and combine them to produce synthetic representation. Synthesizing these distinct representation of different samples is crucial, because the optimal classifier should consistently identify target object irrespective of background. The features are separated to be mutually exclusive with $z^o$ and $z^b$. $ z^o$ is highly related to the class label prediction, whereas $z^b$ is correlated with the object, although it is not necessary to predict the class label. In other words, for an image $x$, even if the background representation $z^b$ is replaced with another representation $z^{b*}$, the predictions are assumed to remain unaffected. Therefore, the optimal classifier $f^*$ should provide consistent predictions without being affected by biases. Our hypothesis can be expressed as Eq~\ref{eq:equation4}. 

\begin{equation}
f^{*}([z^o, z^b]) = f^{*}([z^o, z^{b*}])
\label{eq:equation4}
\end{equation}

$[\cdot,\cdot]$ denotes the channel-wise concatenation; that is, to achieve such an invariant prediction, shuffling the separate representation is employed. First, we randomly permute the disentangled background representation and obtain $\hat{z}^b$. Then, we concatenate $\hat{z}^b$ with $z^o$ to create a novel representation $z^{sb}$ represents a fixed object-relevant representation combined with a swapped background representation from other image in the mini-batch. Additionally, to provide a more diverse representation to the classifier, the augmentation is also performed in the opposite direction. That is, $\hat{z}^o$ is obtained by randomly shuffling $z^o$ and then obtaining $z^{so} = [\hat{z}^o, z^b]$ by concatenating with $z^b$. Then, we feed $z^{sb}$ to the classifier $f_s$ and supervise the classification score with the target label $y$. In the case of $z^{so}$, the target object is shuffled in the mini-batch, and thus, the target label y is also rearranged to $\hat{y}$, according to the permuted index.

\begin{equation}
L_{shuffle} = BCE(f_s(z^{sb}), y) + BCE(f_s(z^{so}), \hat{y})
\label{eq:equation5}
\end{equation}

The objective function for training the classifier augmented with the shuffled representation can be expressed as Eq~\ref{eq:equation5}. 

\begin{equation}
L = L_{cls} + \lambda L_{contr} + L_{shuffle}
\label{eq:equation6}
\end{equation}

Thus, the total loss function can be described as Eq~\ref{eq:equation6}, where $\lambda$ represents the balancing scalar. The overview of SMA are shown in Figure~\ref{fig:fig3}. More results on $\lambda$ can be found in supplementary material.

\begin{figure*}[t]
\begin{center}
   \includegraphics[width=1.0\linewidth]{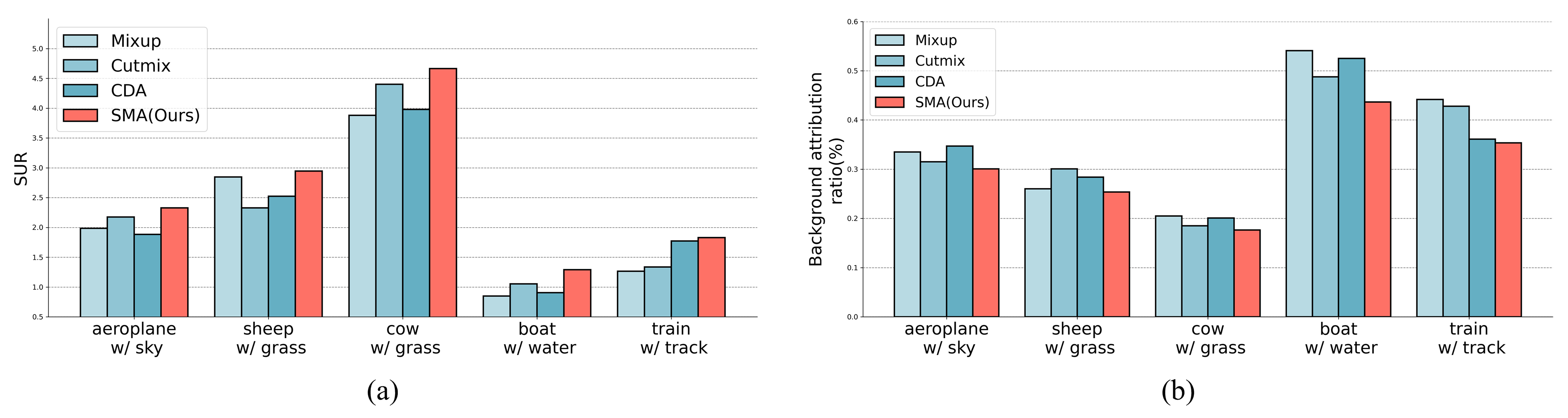}
\end{center}
\vspace*{-6mm}
   \caption{For each object–background pair, we evaluated (a) shortcut usage ratio (SUR) (\textcolor{blue}{higher the better}) and (b) background attribution ratio (\textcolor{blue}{lower the better}) for each augmentation including Mixup\cite{zhang2017mixup}, CutMix\cite{yun2019cutmix}, CDA\cite{su2021context}, and SMA(Ours).}
\label{fig:fig4}
\end{figure*}

\noindent \textbf{Discussions.} Two-way shuffling combines the object-relevant attributes in the representation space with the background attributes that do not appear frequently with the corresponding class. Consequently, the classifier learns the representation that rarely appears in the biased dataset and is thus less dependent on the shortcut features. For example, when images of an ``aeroplane'' with ``sky'' and ``cows and sheeps'' with ``grass landscape'' are used as the input images, the classifier can learn the representations related to ``aeroplane with grass landscape'' and ``cows and sheeps appearing in sky'' in the representation space, which the images of such scenes are absent in the training dataset. Furthermore, the diversity of the representation is guaranteed, because the representations in the mini-batch are randomly combined in every iteration. 

\subsection{Analysis of the shortcut behaviors}
\label{analysis}

In this section, we designed a toy experiment to analyze the shortcut learning behavior of the classifier. First, we collected aeroplane–sky, sheep–grass, cow–grass, boat–water, and train–track images, which are the most frequent object–background pairs in the Pascal VOC 2012 dataset\cite{everingham2010pascal}. The Pascal CONTEXT\cite{mottaghi2014role} dataset was used to determine whether the target background was included. Then, ResNet-50\cite{he2015deep} pretrained with ImageNet\cite{deng2009imagenet} was trained to predict the class labels with the Pascal VOC 2012 dataset after applying different augmentation methods, i.e., Mixup\cite{zhang2017mixup} and CutMix\cite{yun2019cutmix}, which are popular augmentation methods used in classification tasks, as well as CDA\cite{su2021context}, which is applied to WSSS tasks.

The degree of shortcut usage by the classifier can be measured by the contribution of object region ($R_o$) pixels compared to background region ($R_b$) pixels in predicting class labels. To quantify this, we leverage Integrated Gradients(IG) \cite{sundararajan2017axiomatic}, which is a gradient-based method that integrates $m$ gradients along a straight-line path from a baseline $x_{base}$ to an input $x$. Here, we use a black image with the same resolution as the input image $x$ as the baseline input $x_{base}$ \cite{sundararajan2017axiomatic}. The IG of the $i$-th pixel of input image $x$ can be expressed as follows: 

\begin{equation}
I(x_i) = (x_i - x_{base}) \cdot \sum_{k=1}^m {{\partial {f(x_{base} + {{k} \over {m}} (x_i - x_{base}))}} \over {\partial x_i}} \cdot {1 \over m}
\label{eq:equation7}
\end{equation}

We define the \textbf{shortcut usage ratio (SUR)} of an image as the ratio of IG in the background region ($R_b$) to that of the object region ($R_o$). SUR indicates the extent to which the information from the object region is utilized compared to the background region. SUR can be calculated as follows:

\begin{equation}
SUR(x) = \cfrac{\sum_{i \in R_o} I(x_i)}{\sum_{i \in R_b} I(x_i)}
\label{eq:equation8}
\end{equation}

In addition, to directly evaluate the extent of mitigated shortcut usage, we calculated the \textbf{background attribution ratio (BAR)} between the attributions in the background region and the sum of all attributions when predicting the target class. Shortcut is an unintended decision rule when making predictions, and in our case, the background represents this. Therefore, it is possible to measure the extent to which a classifier uses shortcuts by 1) object-relevant attributions and 2) background attributions when predicting class labels, which can be measured by SUR and BAR respectively.

We computed the average SUR and BAR for all samples of each object–background pair, which can be seen in Figure~\ref{fig:fig4} (a) and (b), respectively. Experimental results show that SMA accomplished highest SUR in all cases of object–background pairs compared to the existing augmentation method. Also, our proposed method achieved a lowest background attribution ratio among the existing methods in all cases. It can be confirmed that the classifier trained through SMA is less dependent on shortcut features and leverage object relevant features more when making predictions.

\subsection{Training schemes}
\label{schemes}

We used several training schemes to improve the performance of the classifier. First, our proposed augmentation method was applied after a certain epochs. We assumed that the two representations are sufficiently separated, before implementing shuffling. However, if the synthesizing is performed when the features are still entangled, then the classifier will be trained with the wrong signals. The classifier preferentially learns the bias-aligned sample (e.g., boat with water) at the beginning of the learning process and learns the bias conflicting sample (e.g., boat with railroad) later~\cite{nam2020learning}. In other words, it is difficult for the classifier to distinguish the target-object-related features from the bias in the early stages of learning. Therefore, the augmentation is implemented after a specific epochs $t_{aug}$, in which the two features are sufficiently disentangled.

In addition, we do not update the concatenated background features based on the shuffle loss $\mathcal{L}_{shuffle}$. The shuffled feature is supervised by the target labels $y$ and $\hat{y}$, and if the background representation is affected by such a supervision, then the classifier can be wrongly updated. Therefore, to prevent this unintended representation learning, the corresponding feature is detached, so that the shuffle loss does not backpropagate to the background features $z^b$ and $z^{sb}$.


\begin{table}
\begin{center}
\renewcommand{\arraystretch}{1.1}
\renewcommand{\tabcolsep}{6.5mm}
\begin{tabular}{ccc} \hline

Method                    & Seed       & Mask \\ \hline \hline
PSA\cite{ahn2018learning}\tiny{CVPR'18} & 48.0       & 61.0  \\
+ SMA (Ours)            & \textbf{51.4}   & \textbf{64.1}  \\ \hline
IRN\cite{ahn2019weakly}\tiny{CVPR'19}   & 48.3       & 66.3  \\
+ SMA (Ours)            & \textbf{52.4}   & \textbf{68.6}  \\ \hline
AdvCAM\cite{lee2021anti}\tiny{CVPR'21}  & 55.6       & 69.9  \\
+ SMA (Ours)            & \textbf{57.8}   & \textbf{70.4}  \\ \hline
AMN\cite{lee2022threshold}\tiny{CVPR'22} & 62.1       & 72.2  \\
+ SMA (Ours)         & \textbf{64.4}   & \textbf{72.7}   \\ \hline 

\end{tabular}
\end{center}
\vspace*{-3mm}
\caption{We evaluate the mIoU (\%) of localization map (Seed) and pseudo-mask (Mask) on various WSSS methods when SMA was applied on Pascal VOC 2012 \textit{train} set.}
\label{tab:table1}
\end{table}


\section{Experiment}

\subsection{Experimental setup}

\noindent \textbf{Dataset and evaluation metric.} We used the Pascal VOC 2012 \cite{everingham2010pascal} dataset with a total of 21 classes (20 object categories and 1 background) and 10,582 image-level class label images augmented by \cite{hariharan2011semantic} for the training. For the evaluation and testing, 1,449 and 1,456 pixel-level labels were used. We also performed training and evaluation on the MS COCO 2014\cite{caesar2018coco} dataset. The MS COCO 2014 dataset consists of 81 classes, including background, with 82,783 and 40,504 images for training and validation, respectively. We analyzed the performances of the generated localization map, pseudo-mask, and semantic segmentation benchmark using the mIoU metric, which is generally used for evaluating segmentation results. 

\begin{table}
\begin{center}
\renewcommand{\arraystretch}{1.1}
\renewcommand{\tabcolsep}{6.5mm}
\begin{tabular}{ccc} \hline
Method                    & \textit{val}      & \textit{test} \\ \hline \hline
PSA\cite{ahn2018learning}\tiny{CVPR'18} & 61.7       & 63.7  \\
+ SMA (Ours)             & \textbf{65.9}   & \textbf{66.8}  \\ \hline
IRN\cite{ahn2019weakly}\tiny{CVPR'19}   & 63.5       & 64.8  \\
+ SMA (Ours)             & \textbf{68.6}   & \textbf{68.7}  \\ \hline
AMN\cite{lee2022threshold}\tiny{CVPR'22}   & 69.5       & 69.6  \\
+ SMA (Ours)             & \textbf{70.9}   & \textbf{70.8}  \\ \hline
\end{tabular}
\end{center}
\vspace*{-3mm}
\caption{We show the results of training the semantic segmentation network with the pseudo-mask obtained by applying SMA to the existing WSSS methods for the Pascal VOC 2012 \textit{val} and \textit{test} set.}
\label{tab:table2}
\end{table}

\noindent \textbf{Implementation details.} We adopted ResNet-50\cite{he2015deep} pretrained on ImageNet\cite{deng2009imagenet} as the clasifier, and the stochastic gradient descent was used as the optimizer. The learning rate was set to 0.1 at the beginning of the training, and then decreased for every iteration by using a polynomial decay function. The classifier was trained for 10 epochs, and the point at which the augmentation was applied, $t_{aug}$ was set to 6 through extensive experiments. The additional results of $t_{aug}$ are found in supplementary material.

\begin{figure*}[t]
\begin{center}
   \includegraphics[width=0.99\linewidth]{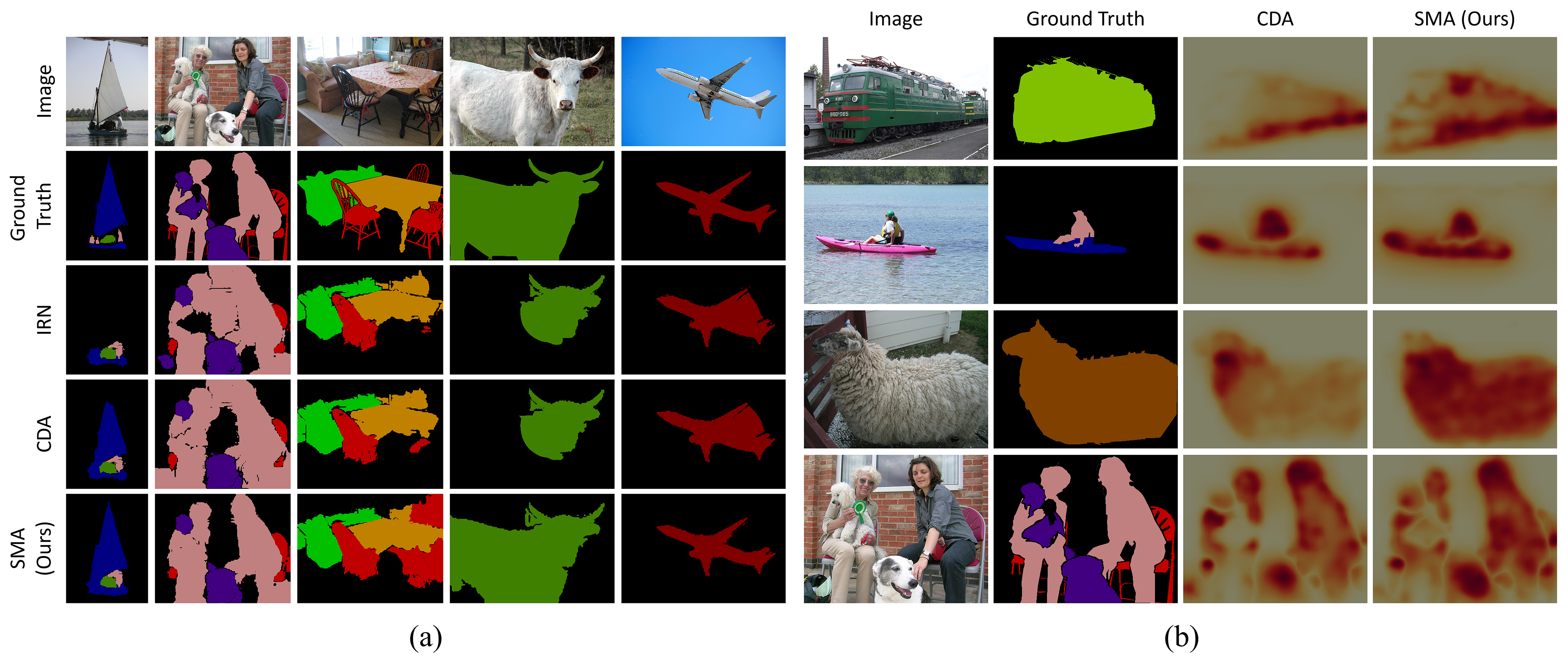}
\end{center}
\vspace*{-6mm}
   \caption{Qualitative examples of (a) pseudo-masks from IRN\cite{ahn2019weakly}, CDA\cite{su2021context} and SMA(Ours) on PASCAL VOC 2012 \textit{val} images (b) localization maps which are obtained by applying CDA\cite{su2021context} and SMA (Ours).}
\label{fig:fig6}
\end{figure*}



\subsection{Experimental results}
\noindent \textbf{Quantitative results of localization maps and pseudo-masks.} Table~\ref{tab:table1} compares the initial seeds (Seed) and pseudo-masks (Mask) obtained by using the WSSS baseline and our proposed methods on Pascal VOC 2012 dataset. When our method was applied to different baselines\cite{ahn2018learning, ahn2019weakly, lee2022threshold}, their performances showed considerable improvements. In particular, when we applied our method on AMN\cite{lee2022threshold}, an mIoU value of 72.7\% was achieved for the pseudo-mask, and this value exceeded those obtained using the previous methods. Compared to other methods, SMA focuses more on the target object while exploiting less background cues. Figure~\ref{fig:fig6} (a) hows examples of the pseudo-masks generated through the CDA\cite{su2021context}, IRN\cite{ahn2019weakly}, and proposed method. These examples also show that compared to the CDA, our method captures the target object more accurately and does not mis-assign the background region to the foreground region.

\noindent \textbf{Quantitative results for the semantic segmentation network.} Table~\ref{tab:table2} shows the results of training the semantic segmentation network with the pseudo-mask obtained by applying the proposed augmentation method to PSA\cite{ahn2018learning}, IRN\cite{ahn2019weakly}, and AMN\cite{lee2022threshold} and the evaluated mIoU value for the Pascal VOC 2012 \textit{val} and \textit{test} sets. As a result of applying SMA, PSA improved by 4.2\%p and 3.1\%p, and IRN improved by 5.1\%p and 3.9\%p for the \textit{val} and \textit{test} sets. In addition, when our proposed method was employed on AMN, mIoU values of 70.9\% and 70.8\% were achieved for the \textit{val} and \textit{test} sets, respectively. These observations confirm that our proposed method can be effectively applied to the previous WSSS methods and can notably boost their semantic segmentation performance.We adopted the MS COCO 2014 dataset, which is a large-scale benchmark with a large number of classes to demonstrate the generalization ability and robustness of our proposed method. Table~\ref{tab:table3} shows the results for the MS COCO validation set. When SMA was employed on AMN, it achieves an improvement of 1.2\%p compared with AMN, and outperforms the other competitive methods. Qualitative results for MS COCO 2014 are in supplementary material.

\noindent \textbf{Comparison with other augmentation methods.} Table~\ref{tab:table4} compares the performance of the localization maps obtained by applying different augmentations including Manifold mixup\cite{verma2019manifold}, Cutout\cite{devries2017improved}, Mixup\cite{zhang2017mixup}, CutMix\cite{yun2019cutmix}, and CDA\cite{su2021context}. The experimental results show that compared with the other augmentations, including the ones used in the image classification and WSSS tasks, our proposed method showed the highest mIoU value. Evidently, when classical augmentation was used, the classifier becomes confused between the object-relevant and background cue. In addition, the result was 1.6\%p higher than that obtained using the CDA\cite{su2021context}. Figure~\ref{fig:fig6} (b) shows localization maps generated by CDA and our proposed method. 

\subsection{Analysis and discussions}

\begin{table}
\begin{center}
\renewcommand{\arraystretch}{1.1}
\renewcommand{\tabcolsep}{6.5mm}
\begin{tabular}{cc} \hline
Method         & \textit{val} \\ \hline \hline
ADL\cite{verma2019manifold}\tiny{TPAMI'20}  & 30.8 \\
CONTA\cite{verma2019manifold}\tiny{NeurIPS'20}  & 33.4 \\
CDA\cite{yun2019cutmix}\tiny{ICCV'21}              & 33.7 \\
IRN\cite{devries2017improved}\tiny{CVPR'19}    & 41.4 \\
RIB\cite{zhang2017mixup}\tiny{CVPR'21}          & 43.8 \\ 
AMN\cite{lee2022threshold}\tiny{CVPR'22}              & 44.7 \\ \hline
\textbf{AMN\cite{lee2022threshold} + SMA (Ours)}      & \textbf{45.9} \\ \hline
\end{tabular}
\end{center}
\vspace*{-3mm}
\caption{We evaluated semantic segmentation results (mIoU) on MS COCO 2014 \textit{val} set.}
\label{tab:table3}
\end{table}

\begin{table}
\begin{center}
\renewcommand{\arraystretch}{1.1}
\renewcommand{\tabcolsep}{6.5mm}
\begin{tabular}{cc} \hline
Method         & Seed \\ \hline \hline
w/o augmentation                                     & 48.3 \\
Manifold mixup\cite{verma2019manifold}\tiny{ICML'19} & 48.7 \\
Cutout\cite{devries2017improved}\tiny{preprint'17}       & 48.9 \\
Mixup\cite{zhang2017mixup}\tiny{ICLR'18}             & 49.0 \\ 
CutMix\cite{yun2019cutmix}\tiny{ICCV'19}             & 49.2 \\
CDA\cite{su2021context}\tiny{ICCV'21}                & 50.8 \\ \hline
\textbf{SMA (Ours)}                             & \textbf{52.4} \\ \hline
\end{tabular}
\end{center}
\vspace*{-3mm}
\caption{We compare localization map performance of various augmentation methods on Pascal VOC 2012 \textit{train} set.}
\label{tab:table4}
\end{table}

\begin{table}
\begin{center}
\renewcommand{\arraystretch}{1.1}
\renewcommand{\tabcolsep}{3.0mm}
        \begin{tabular}{ccccccc} \hline
Loss                          & (a)                     & (b)        & (c)        & (d)        & (e)         & Seed  \\ \hline \hline
baseline                      & \checkmark              &            &            &            &             & 48.3  \\
$\mathcal{L}_{cls}$           & \checkmark              & \checkmark &            &            &             & 49.2 \\
$\mathcal{L}_{contr}$         & \checkmark              & \checkmark & \checkmark &            &             & 49.8 \\
$\mathcal{L}_{shuffle}^\dagger$  & \checkmark              & \checkmark & \checkmark & \checkmark &             & 50.9 \\
$\mathcal{L}_{shuffle}^\ddagger$ & \checkmark              & \checkmark & \checkmark & \checkmark & \checkmark  & \textbf{52.4} \\ \hline
\end{tabular}
\end{center}
\vspace*{-3mm}
\caption{Effectiveness of each loss term on localization map on Pascal VOC 2012 \textit{train} set.}
\label{tab:table5}
\end{table}

\begin{table}[]
\begin{center}
\begin{tabular}{cccc}
\hline
\multirow{2}{*}{\begin{tabular}[c]{@{}c@{}}Class\\ (w/ background)\end{tabular}} & \multirow{2}{*}{\begin{tabular}[c]{@{}c@{}}Co-occurrence\\ ratio\end{tabular}} & \multicolumn{2}{c}{Method} \\
                                                                                 &                                                                                & IRN\cite{ahn2019weakly}          & SMA         \\ \hline \hline
\begin{tabular}[c]{@{}c@{}}aeroplane\\ (w/ sky)\end{tabular}                     & 0.23                                                                           & 83.72        & \textbf{87.34}       \\ \hline
\begin{tabular}[c]{@{}c@{}}sheep\\ (w/ grass)\end{tabular}                       & 0.22                                                                           & 85.95        & \textbf{86.10}       \\ \hline
\begin{tabular}[c]{@{}c@{}}cow\\ (w/ grass)\end{tabular}                         & 0.20                                                                           & 86.24        & \textbf{87.79}       \\ \hline
\begin{tabular}[c]{@{}c@{}}boat\\ (w/ water)\end{tabular}                        & 0.18                                                                           & 75.05        & \textbf{76.36}       \\ \hline
\begin{tabular}[c]{@{}c@{}}train\\ (w/ track)\end{tabular}                       & 0.11                                                                           & 68.83        & \textbf{71.69}      \\ \hline 
\end{tabular}
\end{center}
\vspace*{-3mm}
\caption{We evaluated mIoU (\%) of localization map for images of object–background pairs that frequently appear in the Pascal VOC 2012 training dataset to analyze spurious correlations.}
\label{tab:table6}
\end{table}

\noindent \textbf{Ablations for loss terms on localization maps.} We performed an ablation study for each loss term as shown in Table~\ref{tab:table5}, where baseline denotes the baseline loss without the proposed augmentation method. $\mathcal{L}_{shuffle}^\dagger$ and $\mathcal{L}_{shuffle}^\ddagger$ are the first and second terms of $\mathcal{L}_{shuffle}$ and, indicate the losses corresponding to the background and object relevant synthesizing, respectively. When the features were disentangled by adding $\mathcal{L}_{cls}$ and $\mathcal{L}_{contr}$, the performance improved by 1.5\%p compared to that obtained using the baseline, implying that the classifier focuses more on the target object region. As a result of randomly shuffling the background representation by adding $\mathcal{L}_{shuffle}^\dagger$, the performance increased by 1.1\%p, compared to that of $\mathcal{L}_{contr}$. Furthermore, when object relevant representation was shuffled by adding the $\mathcal{L}_{shuffle}^\dagger$ term, an mIoU of 52.4\% was achieved. These experimental results show that SMA effectively diversifies the representations and suppress the use of the shortcut features.

\noindent \textbf{Analysis for spurious correlations.} The mIoU values presented in Table~\ref{tab:table6} were obtained for bias-aligned images, which corresponds to frequently appearing object classes and background pairs. We used images with the same object–background combination used in section \ref{analysis}. The co-occurrence ratio denotes the ratio in which the corresponding background appears among the images corresponding to a specific class label. For example, in the entire aeroplane image, the sky coincides by more than 23\%. The removal of harmful correlations was verified by evaluating the performance of the images of the above-mentioned combinations. All the combinations showed results that were higher than those obtained using the IRN\cite{ahn2019weakly}. Thus, it can be confirmed that our proposed method can effectively alleviate the high usage of shortcuts caused by a biased dataset. A more experiment on bias-conflicting samples is provided in supplementary material.

\begin{figure}[t]
\begin{center}
   \includegraphics[width=0.87\linewidth]{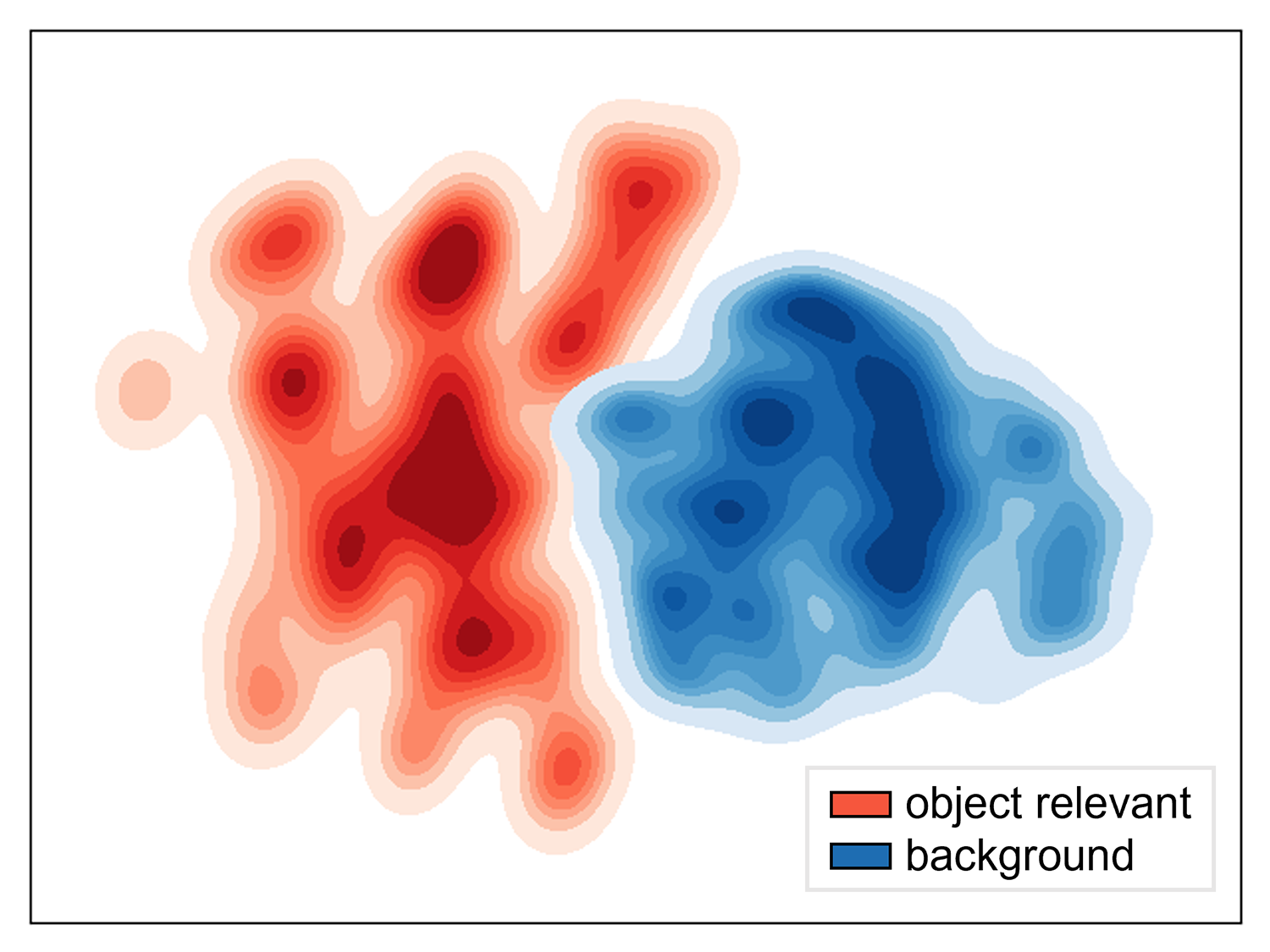}
\end{center}
\vspace*{-5mm}
   \caption{Visualization of the features corresponding to the object relevant and background by using the T-SNE\cite{van2008visualizing} dimension reduction method.}
\label{fig:fig7}
\end{figure}

\noindent \textbf{Manifold visualization.} We performed manifold visualization using T-SNE\cite{van2008visualizing}, which is a dimensionality reduction method, to evaluate the utility of the disentangled object-relevant and background representations. We used the intermediate features $z^o$ and $z^b$ obtained using the aggregators $M_{o}$ and $M_{b}$. Figure~\ref{fig:fig7} reveals that the features related to the foreground and background are semantically different.



\section{Conclusion}

In this study, we propose SMA that disentangles the object-relevant and background representations and shuffles them in mini-batches to create synthetic feature. Evidently, the classifier trained with SMA effectively mitigates the use of shortcuts by focusing on the object itself to perform the predictions. Thus, the performance of the localization map and pseudo-mask obtained through the classifier could be improved by applying our proposed augmentation method. Our method improved localization map and pseudo-mask obtained from a classifier and exhibit increased semantic segmentation performance on Pascal VOC 2012 and MS COCO 2014 datasets. In the future, we intend to combine representations based on distance to allow the classifier to leverage more information related to objects for improving the performance of localization maps.

\noindent \textbf{Limitations.} Our proposed SMA involves a two-step process, which requires additional training interations. Furthermore, while SMA to some extent separates the two representations, it faces challenges in effectively decoupling hard attributes. Despite these limitations, we effectively trained the classifier to detour shortcuts and focus on the target.

\section{Acknowledgements}
This research was supported by Basic Science Research Program through the National Research Foundation of Korea(NRF) funded by the Ministry of Education(NRF-2022R1C1C1008534), and Institute for Information \& communications Technology Planning \& Evaluation (IITP) through the Korea government (MSIT) under Grant No. 2021-0-01341 (Artificial Intelligence Graduate School Program, Chung-Ang University).

{\small
\bibliographystyle{ieee_fullname}
\bibliography{egbib}
}

\pagebreak


\appendix



\counterwithin{figure}{section}

\setcounter{equation}{0}
\renewcommand{\theequation}{A.\arabic{equation}}

\setcounter{table}{0}
\renewcommand{\thetable}{A.\arabic{table}}

\setcounter{figure}{0}
\renewcommand{\thefigure}{B.\arabic{figure}}

This appendix provides further explanation of the additional experiments, and visualizations. Section~\ref{a} describes the ablative results for the hyperparameters and additional experiments. The images selected for the experiments and additional qualitative results are presented in section~\ref{b}.

\section{Additional Experiments}
\label{a}

\noindent \textbf{Hyperparameter analysis.} We analyze the effect of the hyperparameters $t_{aug}$ and $\lambda$ on the initial seed. $t_{aug}$ is the point in the training epoch at which the disentangled representation is shuffled and the ablative results can be seen in Table~\ref{tab:supp_table1}. The experimental results show that suboptimal representations are synthesized when shuffled in a state that is not sufficiently disentangled, and no significant performance improvement is observed even after further training. 

$\lambda$ is a balancing scalar of contrastive loss that determines the degree of separation between object-relevant features and background features.  Table~\ref{tab:supp_table2} shows the results of initial seeds obtained by varying the value of $\lambda$. The experimental results confirm that $\lambda=0.5$ achieved the highest mIoU. 


\noindent \textbf{Experimental results for bias-conflicting samples.} A classifier that exploits the shortcut feature to predict class labels leverages the background attribute and is less dependent on target object related attributes.
Therefore, the classifier fails to identify the target object for samples outside the general context. To analyze this, we selected images that did not contain a top-3 background, which is a background with a high co-occurring ratio with a specific class. That is, a bias-conflicting sample\cite{lee2021learning}, which is an object-background combination that does not often appear in the training dataset, was selected. We evaluated the localization map generated by IRN\cite{ahn2019weakly} and our method; the results of these evaluations can be seen in Table~\ref{tab:supp_table3}. The proposed method produced better results for bias-conflicting samples of the selected class, was less affected by the background, and more accurately captured the target object.

\noindent \textbf{Experiments on interpolation-based feature shuffling.}  We experimented with different combining strategies when synthesizing disentangled features. Inspired by previous mix-based methods\cite{zhang2017mixup, zhou2021mixstyle}, novel representations are produced by interpolating separated features between instances. The representations $z_i$ and $z_j$ are obtained by concating the object-relevant($z^o_i, z^o_j$) and the features($z^b_i, z^b_j$) of the i-th and j-th samples, randomly extracted from the training data.

\begin{equation}
 z_i = [z^o_i, z^b_i], z_j = [z^o_j, z^b_j]
\label{eq:supp_equation1}
\end{equation}

We produce novel feature-target pair as follows:

\begin{table}[t]
\begin{center}
\renewcommand{\arraystretch}{1.1}
\renewcommand{\tabcolsep}{6.5mm}
\renewcommand{\table}{B.\arabic{table}}
\begin{tabular}{cc} \hline
$t_{aug}$        & mIoU (\%) \\ \hline \hline
w/o shuffle      & 49.8 \\
2                & 46.4 \\
4                & 51.7 \\
6                & \textbf{52.4} \\ 
8                & 52.0 \\ \hline
\end{tabular}
\end{center}
\vspace{-4mm}
\caption{Experimental result on the relationship between localization map performance and $t_{aug}$.}
\label{tab:supp_table1}
\end{table}

\begin{table}[t]
\begin{center}
\renewcommand{\arraystretch}{1.1}
\renewcommand{\tabcolsep}{6.5mm}
\begin{tabular}{cc} \hline
$\lambda$        & mIoU (\%) \\ \hline \hline
0.1              & 50.9 \\
0.3              & 51.6 \\
0.5              & \textbf{52.4} \\
0.7              & 51.8 \\ 
1                & 51.2 \\ \hline
\end{tabular}
\end{center}
\vspace{-4mm}
\caption{Effect of values of $\lambda$ on localization map performance.}
\label{tab:supp_table2}
\end{table}

\begin{table}[t]
\setlength\tabcolsep{4.5pt}
\begin{tabular}{cccccc}
\hline 
\multirow{2}{*}{Class} & \multicolumn{3}{c}{\multirow{2}{*}{\begin{tabular}[c]{@{}c@{}} 
w/o background\\ (co-occurrence ratio)\end{tabular}}}                                                             & 
\multicolumn{2}{c}{Method} \\
                       & \multicolumn{3}{c}{}                                                                                                                                                           & IRN\cite{ahn2019weakly}     & SMA              \\ \hline \hline
aeroplane              & \begin{tabular}[c]{@{}c@{}}sky\\ (0.23)\end{tabular}   & \begin{tabular}[c]{@{}c@{}}building\\ (0.11)\end{tabular} & \begin{tabular}[c]{@{}c@{}}grass\\ (0.10)\end{tabular}    & 56.45   & \textbf{57.83}   \\
sheep                  & \begin{tabular}[c]{@{}c@{}}grass\\ (0.22)\end{tabular} & \begin{tabular}[c]{@{}c@{}}tree\\ (0.11)\end{tabular}     & \begin{tabular}[c]{@{}c@{}}fence\\ (0.08)\end{tabular}    & 61.38   & \textbf{63.26}   \\
cow                    & \begin{tabular}[c]{@{}c@{}}grass\\ (0.20)\end{tabular} & \begin{tabular}[c]{@{}c@{}}tree\\ (0.14)\end{tabular}     & \begin{tabular}[c]{@{}c@{}}sky\\ (0.13)\end{tabular}      & 60.15   & \textbf{62.36}   \\
boat                   & \begin{tabular}[c]{@{}c@{}}water\\ (0.18)\end{tabular} & \begin{tabular}[c]{@{}c@{}}sky\\ (0.16)\end{tabular}      & \begin{tabular}[c]{@{}c@{}}building\\ (0.10)\end{tabular} & 59.62   & \textbf{63.33}   \\
train                  & \begin{tabular}[c]{@{}c@{}}track\\ (0.11)\end{tabular} & \begin{tabular}[c]{@{}c@{}}ground\\ (0.11)\end{tabular}   & \begin{tabular}[c]{@{}c@{}}sky\\ (0.11)\end{tabular}      & 53.66   & \textbf{58.82}  \\ \hline
\end{tabular}
\vspace{-1mm}
\caption{We selected an image excluding the specified object with frequently appearing top-3 background from the Pascal VOC 2012 dataset and evaluated the mIoU (\%) of the localization map for those samples.}
\label{tab:supp_table3}
\end{table}

\begin{table}[t]
\begin{center}
\renewcommand{\arraystretch}{1.1}
\renewcommand{\tabcolsep}{6.5mm}
\begin{tabular}{cc} \hline
$\alpha$        & mIoU (\%) \\ \hline \hline
0.2              & 51.4 \\
0.4              & 51.3 \\
0.6              & 51.5 \\
0.8              & 51.2 \\ 
1.0              & 51.4 \\ 
\textbf{SMA (Ours)}                & \textbf{52.4} \\ \hline
\end{tabular}
\end{center}
\caption{Performance of localization map (mIoU) generated by interpolation-based method on various $\alpha$ values.}
\vspace{-8mm}
\label{tab:supp_table4}
\end{table}

\begin{figure*}[t!]
\begin{center}
   \includegraphics[width=0.85\linewidth]{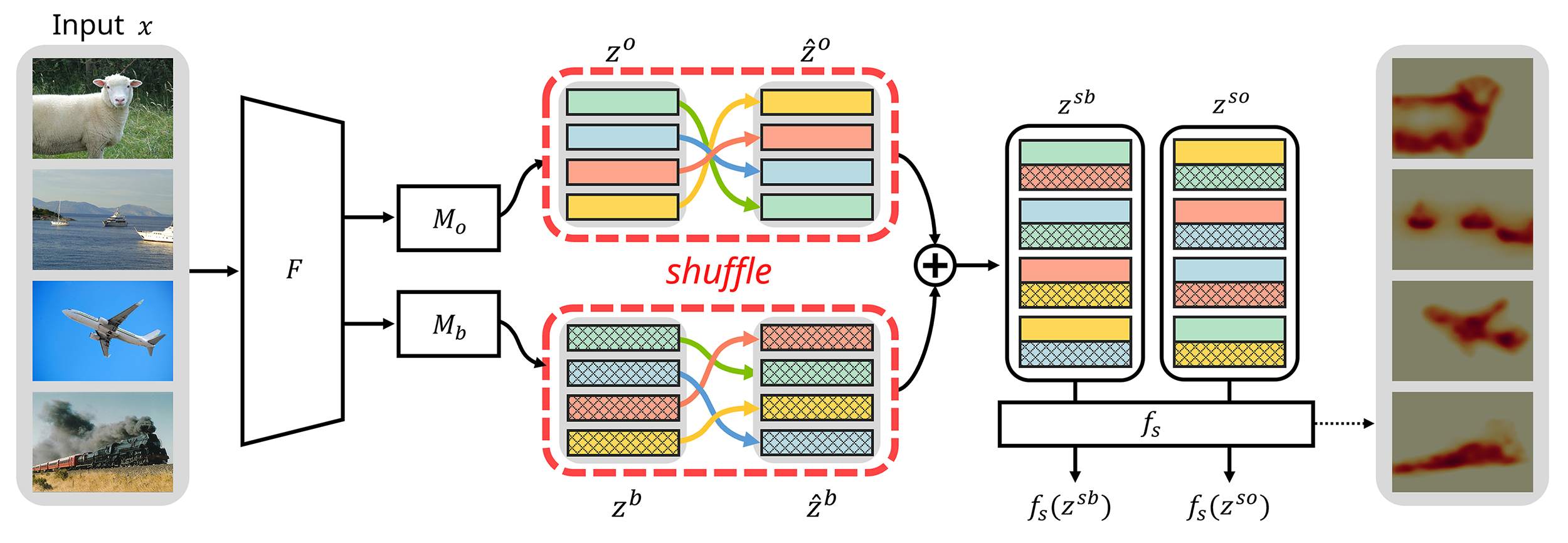}
\end{center}
   \caption{A detailed overview of SMA when representations are shuffled and combined in a two-way manner.}
\label{fig:supp_fig1}
\end{figure*}

\begin{equation}
\tilde{z} = \delta z_i + (1 - \delta) z_j,\; \tilde{y} =  \delta y_i + (1 - \delta) y_j,
\label{eq:supp_equation2}
\end{equation}

where $\delta \sim Beta(\alpha, \alpha)$, for $\alpha \in (0, \infty)$. $y_i$ and $y_j$ are the target vector corresponding to $z_i, z_j$, and $\delta \in [0, 1]$. The hyperparameter $\alpha$ determines the strength of the interpolation between two pairs.

We only change the combining strategy, and all other experimental settings and hyperparameters are the same as our proposed method. The quality of localization maps generated by an interpolation-based method is measured by mIoU. Table~\ref{tab:supp_table4} summarizes the results of an interpolation-based method on various $\alpha$ values. As a result of the experiment, it was confirmed that the performance was limited compared to SMA. We hypothesize that the blending of background attributes hinders the classifier's ability to accurately distinguish the foreground.

\section{More Visualizations}
\label{b}

\setcounter{figure}{1}
\renewcommand{\thefigure}{B.\arabic{figure}}

\noindent \textbf{An overview of SMA.} Figure~\ref{fig:supp_fig1} shows the overall process of our proposed SMA.

\begin{figure*}[t]
\begin{center}
   \includegraphics[width=0.85\linewidth]{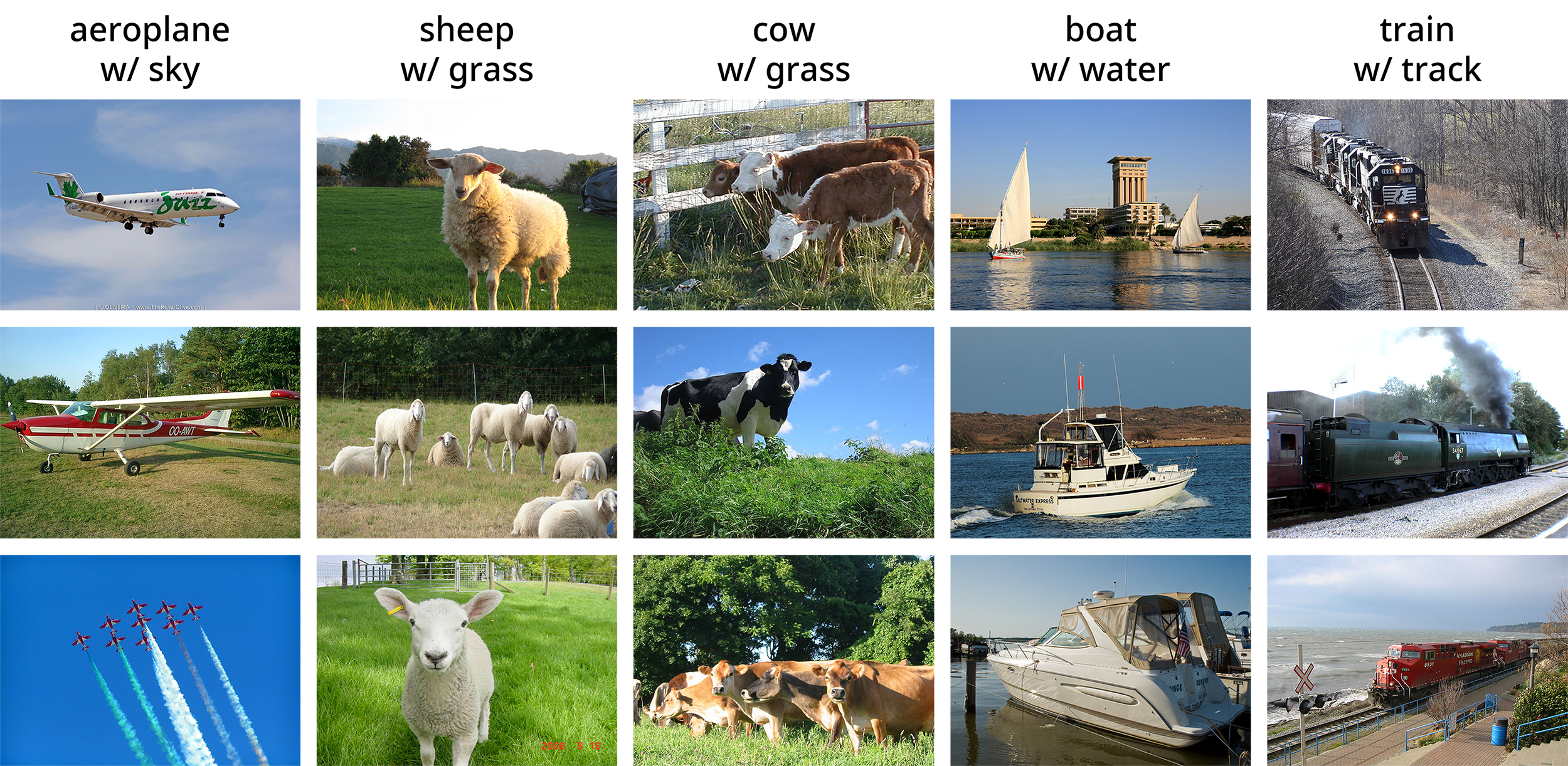}
\end{center}
   \caption{Examples of \textbf{bias-aligned} samples, which refers to images that include a background that frequently appears with a specific object in the Pascal VOC 2012 dataset.}
\label{fig:supp_fig2}
\end{figure*}

\begin{figure*}[t]
\begin{center}
   \includegraphics[width=0.85\linewidth]{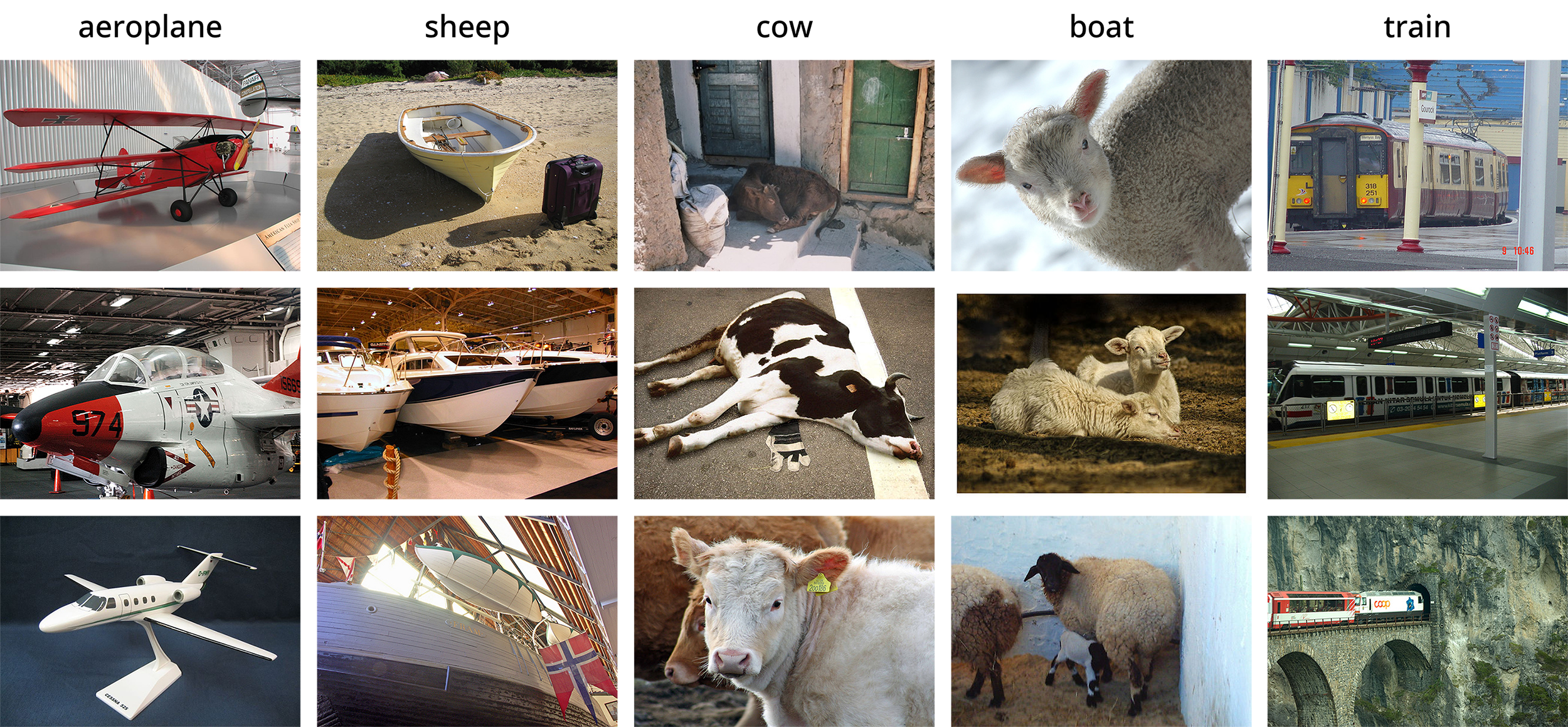}
\end{center}
   \caption{Example of \textbf{bias-conflicting} samples, which is the images that do not include a background that usually appears with a specific object in the Pascal VOC 2012 dataset.}
\label{fig:supp_fig3}
\end{figure*}

\noindent \textbf{Examples of bias-aligned and bias-conflicting samples.} Figure~\ref{fig:supp_fig2} presents examples of bias-aligned samples\cite{nam2020learning} which is the images of the object-background combination with the high co-occurrence frequency. Example images of bias-conflicting samples\cite{lee2021learning}, in which certain objects and frequently appearing backgrounds are intentionally excluded, can be seen in Figure~\ref{fig:supp_fig3}.

\begin{figure*}[t]
\begin{center}
   \includegraphics[width=0.97\linewidth]{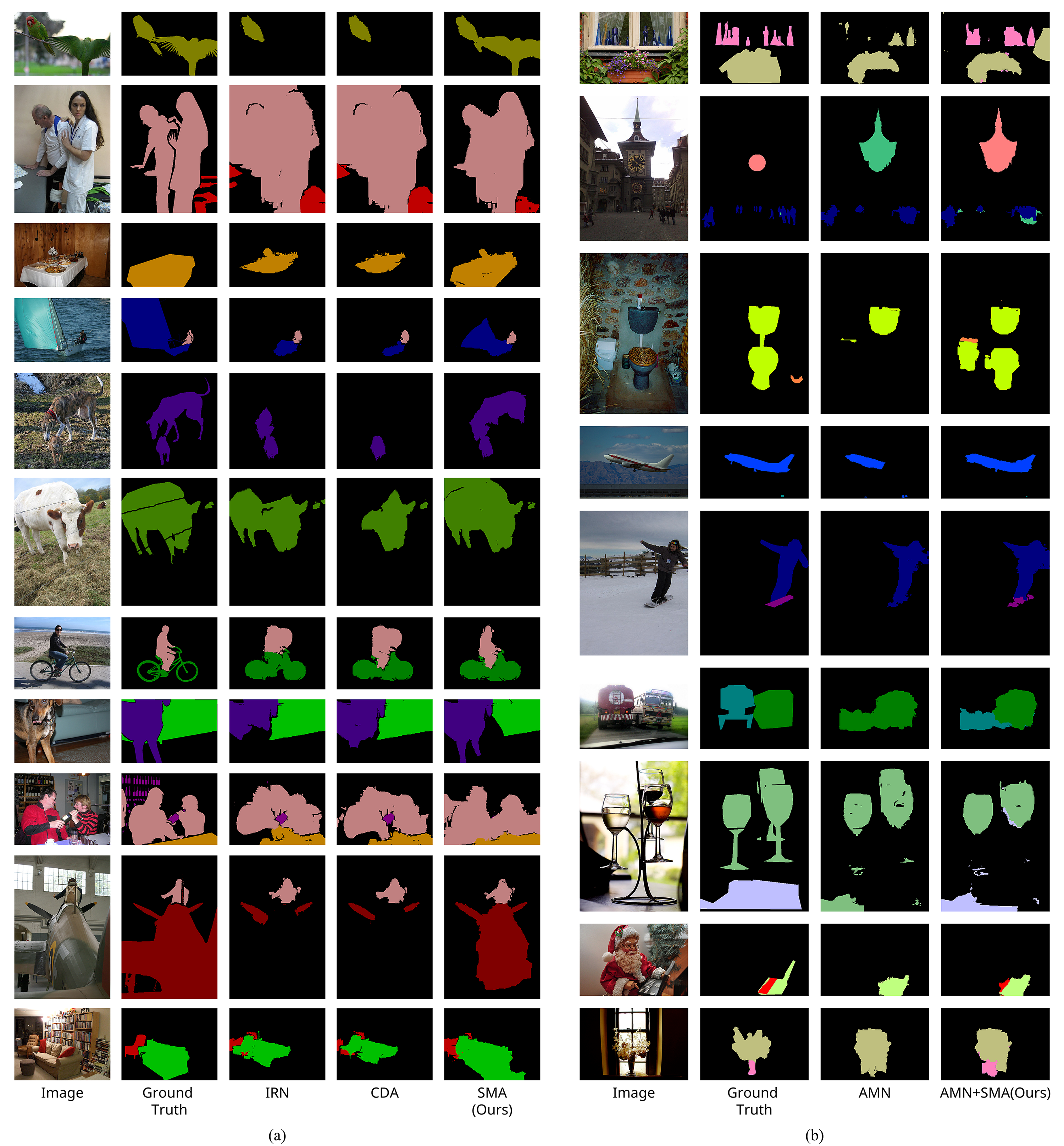}
\end{center}
   \caption{Examples of pseudo-masks from (a) IRN\cite{ahn2019weakly}, CDA\cite{su2021context} and SMA (Ours) on Pascal VOC 2012 (b) AMN\cite{lee2022threshold} and SMA (Ours) on MS COCO 2014 datasets. }
\label{fig:supp_fig4}
\end{figure*}

\noindent \textbf{More qualitative examples.} Figure~\ref{fig:supp_fig4} (a) shows examples of pseudo-masks predicted by IRN\cite{ahn2019weakly}, CDA\cite{su2021context}, and our proposed method SMA on Pascal VOC 2012 dataset, and Figure~\ref{fig:supp_fig4} (b) presents comparison of qualitative results of pseudo-masks produced by AMN\cite{lee2022threshold} and our method on MS COCO 2014 dataset.

\end{document}